\documentclass[letterpaper, 10 pt, conference]{ieeeconf}  % Comment this line out if you need a4paper

\IEEEoverridecommandlockouts                              % This command is only needed if 
\usepackage{graphicx,subfig}

\usepackage[hidelinks]{hyperref}
\usepackage[backend=bibtex,sorting=none,style=ieee]{biblatex}
\usepackage{booktabs}  
\usepackage{threeparttable} 
\usepackage{makecell}
\usepackage{multicol}
\usepackage{multirow}
\usepackage{feyn}
\usepackage{amsmath,amssymb,amsfonts}
\usepackage{ulem}
\usepackage{fancyhdr}
\usepackage{float}

\hypersetup{
	hidelinks,
}

\title{\LARGE \bf
IAF-Net: Illumination-Adaptive Fusion for Low-Light Urban Road Segmentation
}

\author{Bingtao Wang, Daojie Peng, Fulong Ma, Jun Ma, Liang Zhang$^\dagger$
\thanks{$\dagger$ Corresponding author: \texttt{201299800013@sdu.edu.cn}}
\thanks{Bingtao Wang and Liang Zhang are with The Shandong University, wangbt@mail.sdu.edu.cn, 201299800013@sdu.edu.cn }
\thanks{Daojie Peng, Fulong Ma and Jun Ma are with The Hong Kong University of Science and Technology (Guangzhou) (e-mail: \{fmaaf, dpeng108\}@connect.hkust-gz.edu.cn, jun.ma@ust.hk.)}
}

\begin{document}

\maketitle

\thispagestyle{empty}
\pagestyle{empty}

%%%%%%%%%%%%%%%%%%%%%%%%%%%%%%%%%%%%%%%%%%%%%%%%%%%%%%%%%%%%%%%%%%%%%%%%%%%%%%%%
\begin{abstract}
Semantic road segmentation is important for autonomous driving, but existing methods suffer severe performance degradation under low-light conditions. Many existing multi-modal fusion methods do not explicitly adapt to illumination-dependent changes in modality reliability, which can propagate degraded RGB features into the fused representation at night. We propose \textbf{IAF-Net} (Illumination-Adaptive Fusion Network), an end-to-end framework with illumination-adaptive fusion for robust road segmentation across different lighting conditions. It dynamically adjusts fusion weights of RGB and geometric features via the core Illumination-Adaptive Fusion (IAF) module, and enhances low-light feature selection with a brightness-modulated attention decoder. We also construct two dedicated datasets: nuScenes Nighttime Road Segmentation (nuScenes-NRS) and CARLA Multi-Weather Road Segmentation (CARLA-MWRS). Experiments on nuScenes-NRS show state-of-the-art overall performance among the compared methods, while CARLA-MWRS further validates robustness across adverse weather conditions. Ablation studies on a 40\% training subset further highlight the importance of the IAF module, which provides the largest individual gain of 0.70\% in MaxF.

\end{abstract}

\begin{keywords}
Low-light visual perception, sensor fusion, computer vision for transportation, deep learning for visual perception, multimodal semantic perception.
\end{keywords}

\begin{figure*}[t]
    \centering
    \includegraphics[width=1.0\linewidth]{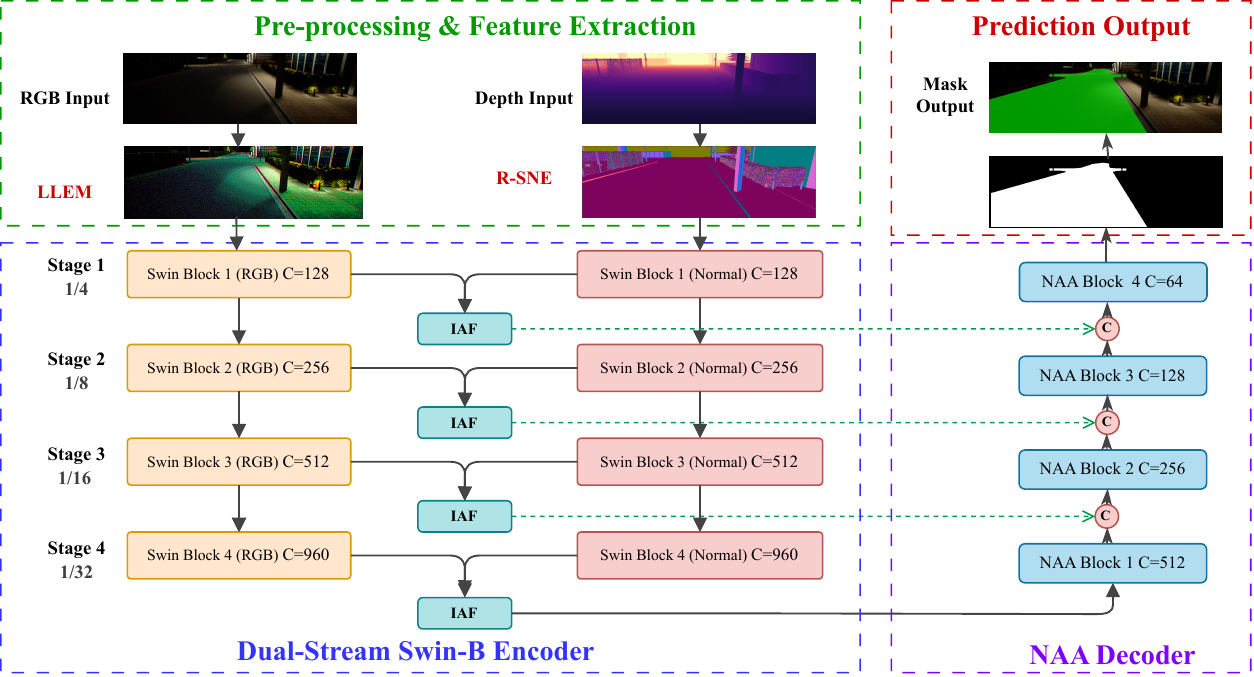}
    \caption{Overall architecture of the proposed IAF-Net. The network takes RGB images and depth maps as inputs, and processes them through four core modules: LLEM for low-light enhancement and brightness estimation, R-SNE for robust surface normal feature extraction, IAF for illumination-adaptive multi-scale feature fusion, and NAA for nighttime-aware attention decoding. An auxiliary EdgeHead is adopted to enhance boundary prediction.}
    \label{fig:architecture}
\end{figure*}

\section{Introduction}
Robust perception under diverse illumination and weather conditions is a key requirement for autonomous driving systems. As a core component of the autonomous driving perception system, semantic road segmentation aims to identify the drivable free space from the scene, providing essential information for path planning, obstacle avoidance, and vehicle control. 
% While significant progress has been made in daytime road segmentation, the performance of existing methods still degrades severely under low-light and nighttime conditions, which account for a large proportion of traffic accidents \cite{nighttime_accident2020}.
Current research on road segmentation spans single-modal methods \cite{sun2025rod,milioto2019rangenet++}, multi-modal fusion approaches \cite{feng2025sneroadsegv2,chen2019progressive,min2022orfd,peng2026litevilnet}, and recent extensions to self-supervised label generation \cite{ma2023self}. However, existing methods still suffer severe performance degradation under low-light and nighttime conditions. The key limitation is not the absence of nighttime perception studies, but the insufficient modeling of illumination-dependent modality reliability in RGB-normal road segmentation, despite the well-documented difficulty of low-light visual perception \cite{ghari2024lowlightsurvey}. This limitation is related to the broader challenge of adverse-condition perception, where nighttime illumination, rain, and fog change the data distribution of segmentation models \cite{zheng2023robustperception}.

The main challenges of nighttime road segmentation stem from complex and dynamic illumination conditions. First, heterogeneous light sources, such as street lamps, vehicle headlights, building lights, and traffic signals, cause severe spatial variation of illumination. Second, low-light RGB images suffer from low brightness, reduced contrast, and sensor noise, which corrupt the texture and color features used by traditional segmentation methods \cite{ghari2024lowlightsurvey,zerodce2020}. Third, many existing multi-modal fusion methods do not explicitly account for illumination-dependent modality reliability \cite{zhang2023cmx,li2024roadformer}, which may propagate degraded RGB noise into the fused features at night.

To address these challenges, we propose IAF-Net, an illumination-adaptive multi-modal fusion network designed specifically for robust low-light urban road segmentation. The key insight of our work is to use global brightness as a scene-level estimate of RGB reliability rather than as a complete description of local illumination. Local degradation is further handled by low-light enhancement and nighttime-aware attention decoding. Specifically, our network first estimates the global brightness of the scene while enhancing the low-light RGB image. Based on this brightness signal, the network dynamically adjusts the fusion weights of RGB texture features and illumination-invariant geometric features extracted from depth maps. In this way, the network can automatically rely more on geometric features in extremely dark scenes where RGB features are unreliable, and leverage more texture information in well-lit scenes where RGB features are informative.

Our main contributions are summarized as follows:
\begin{itemize}
    \item We propose IAF-Net, an end-to-end framework for low-light road segmentation that integrates illumination awareness into the perception pipeline and adapts to varying lighting conditions.
    \item We design the Illumination-Adaptive Fusion (IAF) module, which leverages the global brightness signal to dynamically adjust the fusion weights of multi-modal features, effectively suppressing noise propagation from degraded RGB features in nighttime scenes.
    \item We propose the Nighttime-aware Attention Decoder (NAA), which introduces a brightness-modulated attention sharpening mechanism to enhance feature selection under low-light conditions, improving the discriminability of features in dark environments.
    \item We construct two dedicated datasets for low-light road segmentation: nuScenes-NRS and CARLA-MWRS, covering real-world nighttime and synthetic multi-weather conditions.
    \item Experiments on nuScenes-NRS show state-of-the-art overall performance among the compared methods, while CARLA-MWRS validates robustness across adverse weather conditions and ablation studies quantify the contribution of each module.
\end{itemize}

% \begin{figure*}[t]
%     \centering
%     \includegraphics[width=1.0\linewidth]{figures/fig_architecture_combined.pdf}
%     \caption{(a) Overall architecture of the proposed IAF-Net. The network takes RGB images and depth maps as inputs, and processes them through four core modules: LLEM for low-light enhancement and brightness estimation, R-SNE for robust surface normal feature extraction, IAF for illumination-adaptive multi-scale feature fusion, and NAA for nighttime-aware attention decoding. An auxiliary EdgeHead is adopted to enhance boundary prediction.
%     (b) Structure of the Lightweight Low-light Enhancement Module (LLEM). The module takes raw low-light RGB images as input, estimates pixel-wise enhancement parameters via lightweight depth-wise separable convolutions, and outputs the enhanced image together with the global brightness estimation for subsequent modules.
%     (c) Pipeline of the Robust Surface Normal Estimation (R-SNE) module. The module preprocesses the raw depth map to remove outliers and noise, computes the surface normal, and generates a confidence-weighted normal map for robust multi-modal fusion.
%     (d) Structure of the proposed Illumination-Adaptive Fusion (IAF) module. The module takes the brightness signal as input, and predicts dynamic fusion weights for RGB and normal features, which are used to fuse the multi-modal features in a illumination-aware manner.
%     }
%     \label{fig:architecture}
% \end{figure*}

\section{Related Works}
In this section, we review the literature from three perspectives that are most relevant to this work: road segmentation for autonomous driving, low-light and nighttime road scene understanding, and multi-modal fusion for road parsing.

\subsection{Semantic Segmentation for Autonomous Driving}
Semantic segmentation has been extensively studied for autonomous driving. Representative CNN-based models, such as DeepLabv3+ \cite{deeplabv3plus2018} and PSPNet \cite{pspnet2017}, established strong baselines for scene parsing, while more recent architectures, including SegFormer \cite{xie2021segformer}, PIDNet \cite{xu2023pidnet}, and DDRNet \cite{hong2021ddrnet}, further improved the trade-off between accuracy and efficiency in driving scenes. These methods provide effective segmentation backbones, but they are largely developed and evaluated under well-lit conditions. When illumination becomes poor, the reliability of RGB appearance cues decreases substantially, which limits the robustness of conventional road parsing pipelines.

\subsection{Low-Light and Nighttime Scene Understanding}
Low-light scene understanding has been studied from both image enhancement and direct nighttime perception perspectives. Classical enhancement methods, such as Retinex-based modeling \cite{retinex1971}, and learning-based approaches, including KinD \cite{kind2019}, LLNet \cite{llnet2015}, and Zero-DCE \cite{zerodce2020}, can improve image visibility in dark scenes. However, these methods are mainly optimized for restoration quality rather than segmentation-oriented feature reliability, and the enhanced appearance does not always translate into more stable road parsing.

Beyond enhancement, recent studies have directly addressed nighttime semantic segmentation. Wei et al. \cite{wei2023disentangle} improved nighttime parsing through illumination disentanglement, while other methods explored thermal assistance, day-to-night adaptation, cross-domain distillation, and event-based sensing \cite{heatnet2019,transnightseg2024}. Representative examples include DANNet \cite{dannet2021}, cross-domain correlation distillation \cite{ccdistill2022}, CMDA \cite{cmda2023}, and event-based road segmentation with asynchronous likelihood attention \cite{eventaseg2024}. These studies confirm the difficulty of nighttime perception, but they either remain strongly coupled to appearance restoration or domain transfer in RGB space, or require additional sensing and adaptation pipelines. As a result, they do not directly address how the reliability of complementary modalities should be regulated as illumination changes.

\subsection{Multi-Modal Fusion for Robust Perception}
Multi-modal fusion has become an effective direction for robust perception because complementary modalities can compensate for the weakness of RGB observations under adverse conditions. In general semantic segmentation, methods such as CMX \cite{zhang2023cmx} and CAFuser \cite{broedermann2025cafuser} show that cross-modal interaction and condition-aware fusion can improve robustness in RGB-X perception.

For road segmentation, geometric cues have drawn increasing attention because surface normals derived from depth provide structural information that is much less sensitive to illumination changes. Wang et al. \cite{wang2019raldrivable} investigated RGB-D drivable-area learning with self-supervised label generation to reduce manual annotation for road and anomaly segmentation. Li et al. \cite{li2024roadformer} proposed RoadFormer for RGB-normal road scene parsing, and Huang et al. further extended this line in RoadFormer+ \cite{huang2025roadformerplus} with more advanced heterogeneous fusion. However, although existing fusion methods improve robustness by combining complementary modalities, they rarely treat illumination as an explicit control signal for regulating the relative reliability of RGB and geometric cues in road segmentation. Consequently, the fusion process is often insufficiently matched to the large reliability shift of RGB features between well-lit and low-light scenes.

\section{Method}
\begin{figure}[t]
    \centering
    \includegraphics[width=\linewidth]{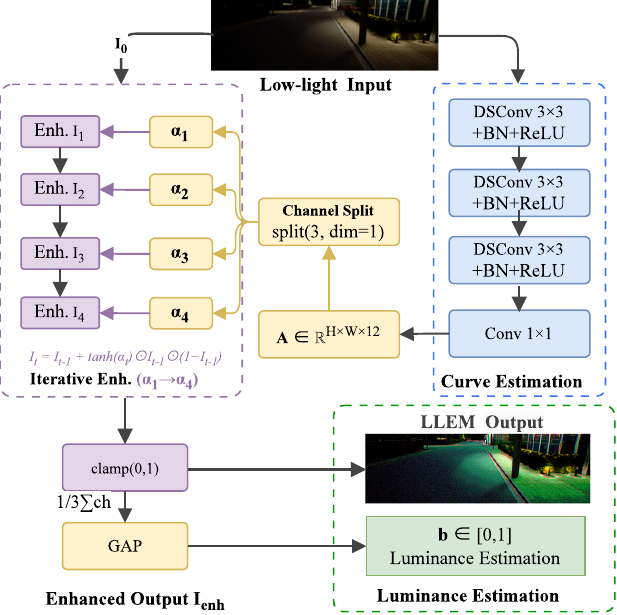}
    \caption{Structure of the Lightweight Low-light Enhancement Module (LLEM). The module takes raw low-light RGB images as input, estimates pixel-wise enhancement parameters via lightweight depth-wise separable convolutions, and outputs the enhanced image together with the global brightness estimation for subsequent modules.}
    \label{fig:llem}
\end{figure}

\begin{figure}[t]
    \centering
    \includegraphics[width=\linewidth]{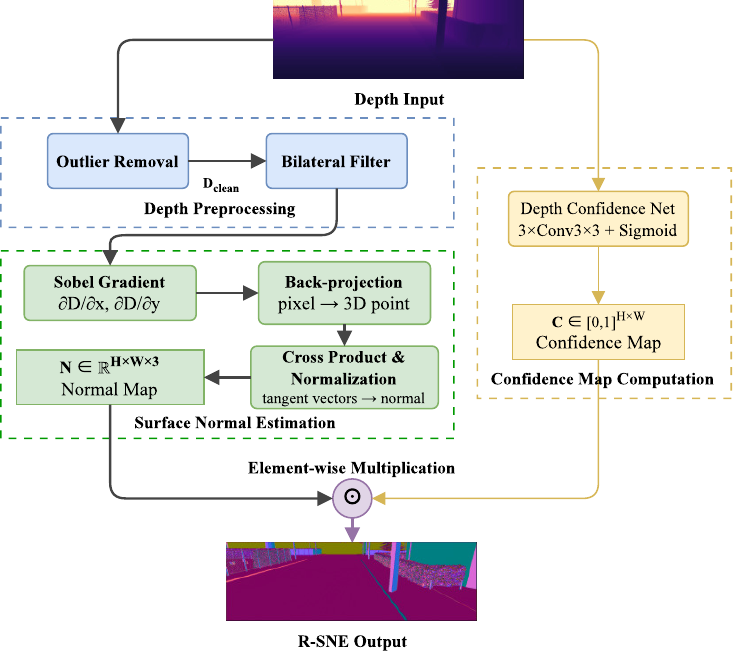}
    \caption{Pipeline of the Robust Surface Normal Estimation (R-SNE) module. The module preprocesses the raw depth map to remove outliers and noise, computes the surface normal, and generates a confidence-weighted normal map for robust multi-modal fusion.}
    \label{fig:rsne}
\end{figure}

In this section, we present the details of the proposed IAF-Net framework. We first introduce the overall architecture, then describe each main module and the associated optimization design.

\subsection{Overall Architecture}
The overall architecture of IAF-Net is illustrated in Fig.~\ref{fig:architecture}. The framework takes an RGB image $I \in \mathbb{R}^{H \times W \times 3}$ and depth-derived geometric input as the two modalities, and processes them through four main modules, followed by an auxiliary boundary branch and brightness-aware loss design, to produce the final road segmentation mask.

First, the raw RGB image is fed into the Lightweight Low-light Enhancement Module (LLEM), which performs adaptive brightness enhancement and outputs the enhanced RGB image as well as a global brightness estimation $b \in [0,1]$ that indicates the overall illumination level of the scene. Meanwhile, the depth signal is converted into surface-normal cues by the Robust Surface Normal Estimation (R-SNE) pipeline, which extracts robust geometric features that are much less sensitive to illumination changes than RGB appearance cues.

Next, the enhanced RGB image and the normal map are fed into a dual-branch Swin Transformer encoder \cite{swin2021}, which extracts hierarchical multi-scale features $\{F_{l}^{rgb}\}_{l=0}^{3}$ and $\{F_{l}^{norm}\}_{l=0}^{3}$ for the two modalities, respectively. These features correspond to resolutions of 1/4, 1/8, 1/16, and 1/32 of the original image size. In the actual training and evaluation pipeline, the normal maps can be pre-computed from depth for efficiency, while the R-SNE module defines the geometric extraction process used by the framework.

Then, the Illumination-Adaptive Fusion (IAF) module takes the multi-scale features from the two modalities and the global brightness signal $b$ as inputs. For each scale, the IAF module dynamically adjusts the fusion weights of RGB and normal features based on the brightness, generating the fused features $\{F_{out}^{(l)}\}_{l=0}^{3}$.

After that, the fused features are fed into the Nighttime-aware Attention Decoder (NAA), which performs hierarchical upsampling and feature integration. The NAA module leverages the brightness signal to modulate the attention mechanism, enhancing the feature selection ability under low-light conditions. Finally, the main segmentation head outputs the road segmentation prediction, while an auxiliary EdgeHead predicts the boundary of the road to further improve the localization accuracy.

\subsection{Low-Light Enhancement and Brightness Estimation}

In nighttime scenes, the raw RGB images are often severely degraded with low brightness and high noise, which hinders feature extraction. To address this, we propose the LLEM module to adaptively enhance low-light images, as shown in Fig.~\ref{fig:llem}. Different from traditional enhancement methods that rely on hand-crafted parameters, LLEM uses a lightweight network with depth-wise separable convolutions to estimate pixel-wise enhancement curve parameters. By iteratively applying the learned enhancement curve, the module improves brightness and contrast while suppressing noise. Different from general low-light enhancement methods that mainly target visual quality \cite{zerodce2020}, LLEM is used to produce both an enhanced image and an explicit brightness cue for downstream fusion and decoding.

In addition to the enhanced image, LLEM also outputs a global brightness estimation $b \in [0,1]$, which measures the overall illumination level of the scene. In practice, $b$ is computed from the enhanced RGB image by first averaging over the three channels and then applying global average pooling to obtain a scalar scene-level brightness estimate for each sample. This brightness signal is then used as a scene-level condition to guide the subsequent fusion, decoding, and loss weighting modules.

\subsection{Robust Surface Normal Estimation}

Surface normal is a geometric feature that describes the orientation of the object surface. Since it is computed from depth information, it is much less sensitive to illumination changes than RGB appearance cues, making it a useful complementary modality for nighttime perception. However, the depth maps captured in nighttime scenes often suffer from noise and missing values, which leads to unstable normal estimation.

To address this, we propose the R-SNE module, which enhances the traditional Surface Normal Estimation (SNE) algorithm with depth preprocessing and confidence estimation. The pipeline of the R-SNE module is shown in Fig.~\ref{fig:rsne}. First, we perform outlier rejection on the raw depth map: for each pixel, we compute the local mean $\mu_{ij}$ and standard deviation $\sigma_{ij}$ in a 5×5 neighborhood, and clean the depth value as shown in Eq.~\eqref{eq:rsne-clean}:
\begin{equation}
\label{eq:rsne-clean}
D_{clean}(i, j)=
\begin{cases}
\mu_{i j}, & \text{if } \dfrac{\left|D(i, j)-\mu_{i j}\right|}{\sigma_{i j}+\epsilon}>\tau_{z}, \\
D(i, j), & \text{otherwise}
\end{cases},
\end{equation}
where $\tau_{z}=3.0$ is the pre-defined threshold for Z-score outlier rejection, and $\epsilon$ is a small positive constant for numerical stability. Then, we apply bilateral filtering to smooth the depth map while preserving edges.

After that, we compute the surface normal from the cleaned depth map. To further improve the reliability, we train a lightweight depth confidence network to predict a pixel-wise confidence map $C \in [0,1]^{H \times W}$, which indicates the reliability of the normal estimation at each pixel. The normal map is then weighted by the confidence map, so that less reliable normal features have lower contribution in the subsequent fusion stage.

\subsection{Illumination-Adaptive Feature Fusion}
\begin{figure}[t]
    \centering
    \includegraphics[width=\linewidth]{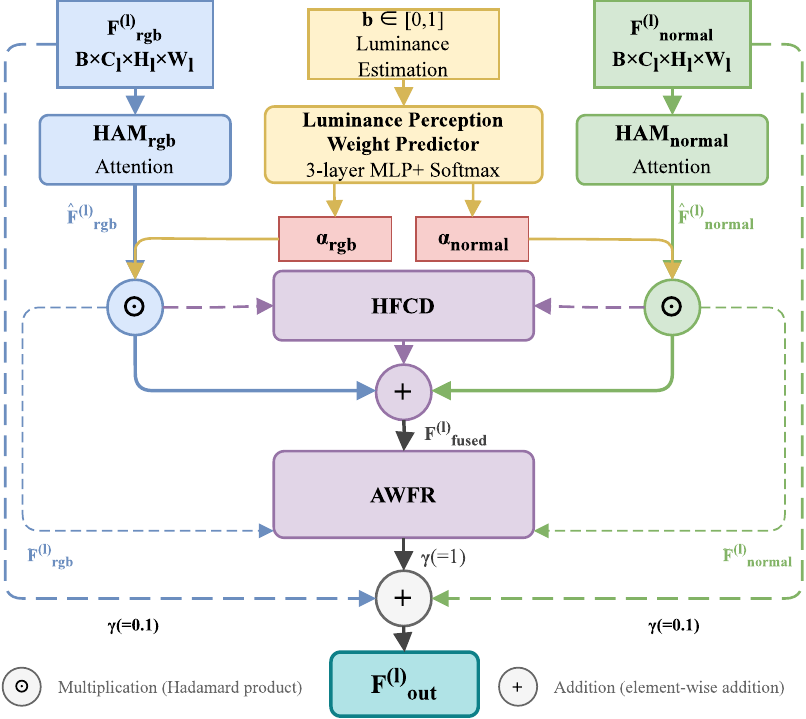}
    \caption{Structure of the proposed Illumination-Adaptive Fusion (IAF) module. The module takes the brightness signal as input, and predicts dynamic fusion weights for RGB and normal features, which are used to fuse the multi-modal features in an illumination-aware manner.}
    \label{fig:iaf}
\end{figure}

The Illumination-Adaptive Fusion (IAF) module is the key component that links the estimated brightness cue to multi-modal feature integration. It addresses the limitation of illumination-agnostic fusion by dynamically adjusting the fusion weights according to the scene brightness. The structure of the IAF module is shown in Fig.~\ref{fig:iaf}.

\begin{figure}
    \centering
    \includegraphics[width=1.0\linewidth]{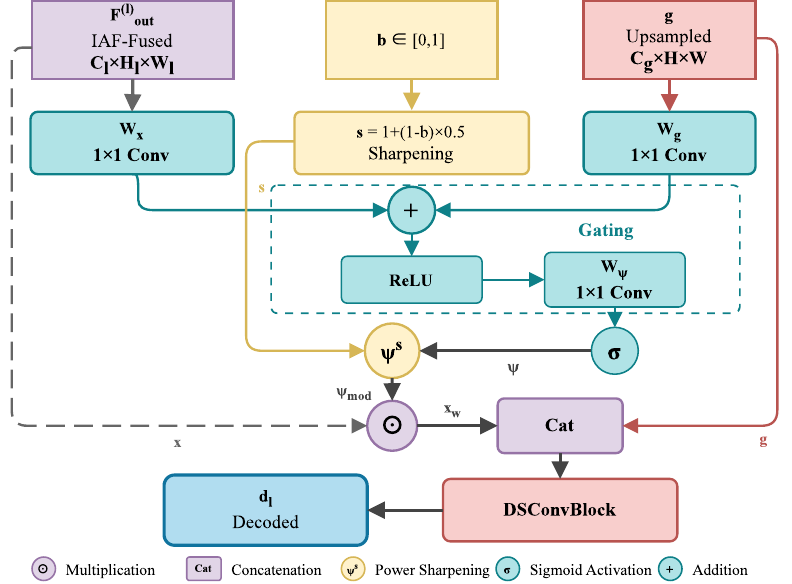}
    \caption{Single-level architecture of the proposed Nighttime-aware Attention Decoder (NAA). The decoder adopts a brightness-modulated attention sharpening mechanism to enhance feature selection under low-light conditions, where the attention map is adaptively sharpened according to the global illumination level.}
    \label{fig:naa}
\end{figure}

The IAF module consists of two main components: a brightness-to-weight MLP predictor and a dynamic feature fusion operation. Given the global brightness signal $b$, the MLP predictor maps this scalar signal into two dynamic fusion weights $\alpha_{rgb}$ and $\alpha_{norm}$, which satisfy $\alpha_{rgb} + \alpha_{norm} = 1$. These weights indicate the relative contribution of RGB and normal features to the fused representation.

Specifically, for each scale $l$, given the RGB features $F_{rgb}^{(l)}$ and normal features $F_{norm}^{(l)}$, we first enhance both modalities with modality-specific hierarchical attention modules (HAM), which reweight intra-modal responses before fusion. The brightness-conditioned weights are then applied to the attended features. A heterogeneous feature contrast descriptor (HFCD) is used to capture complementary cross-modal differences, and an affinity-weighted feature refiner (AWFR) further smooths the fused representation using feature affinity while preserving a residual shortcut from the original two modalities. The fusion stage is formulated as in Eq.~\eqref{eq:iaf-fusion}:
\begin{equation}
\label{eq:iaf-fusion}
\begin{aligned}
F_{out}^{(l)} ={}& \mathrm{AWFR}\!\Big(
\alpha_{rgb}\hat{F}_{rgb}^{(l)} + \alpha_{norm}\hat{F}_{norm}^{(l)} \\
&\quad + \mathrm{HFCD}\!\big(
\alpha_{rgb}\hat{F}_{rgb}^{(l)}, \alpha_{norm}\hat{F}_{norm}^{(l)}
\big)\Big) \\
&\quad + \gamma \left(F_{rgb}^{(l)} + F_{norm}^{(l)}\right).
\end{aligned}
\end{equation}

This design allows the fusion weights to change with the illumination condition. In extremely dark scenes where $b$ is small, the predictor tends to assign a larger $\alpha_{norm}$ and a smaller $\alpha_{rgb}$, so that the fusion result relies more on geometric features and suppresses noise from degraded RGB observations. In better-lit scenes, the predictor can assign a larger $\alpha_{rgb}$ to preserve texture information while still retaining normal cues as geometric support. This behavior is learned end-to-end from the training data.

\begin{figure*}[t]
    \centering
    \includegraphics[width=0.8\linewidth]{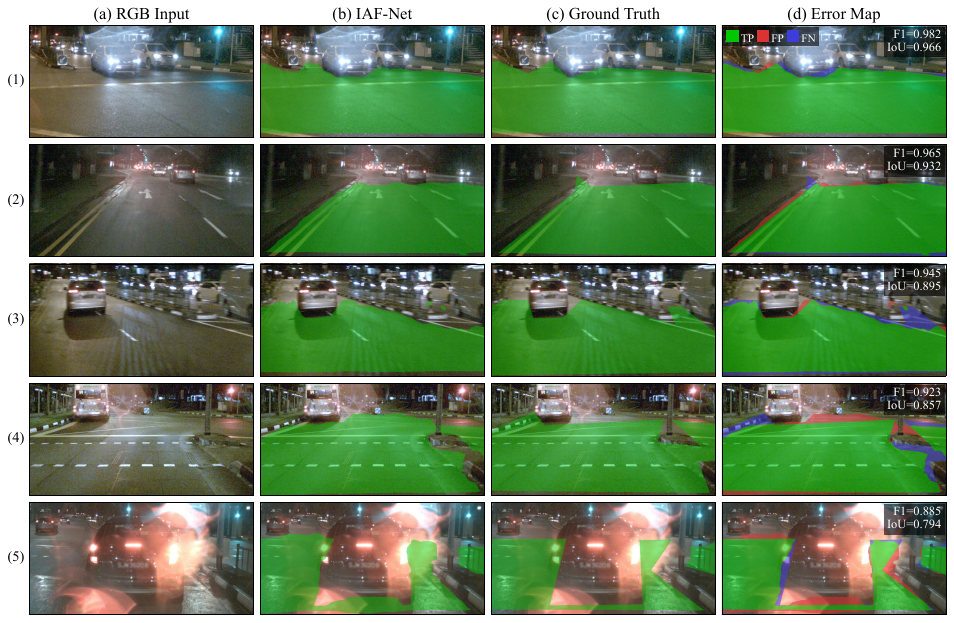}
    \caption{Qualitative segmentation results on nuScenes-NRS. From left to right: input RGB image, our prediction overlay, ground truth overlay, and error map. Green indicates true positives, red indicates false positives, and blue indicates false negatives.}
    \label{fig:qualitative_nuscenes}
\end{figure*}

\subsection{Nighttime-Aware Attention Decoder}

After obtaining the multi-scale fused features, we decode them into the full-resolution segmentation mask. Standard attention-gate mechanisms \cite{oktay2018attention} adopt a fixed attention strength, which is not optimal for nighttime scenes where the noise is more severe and the effective information is sparser.

To address this, we propose the Nighttime-aware Attention Decoder (NAA), whose structure is shown in Fig.~\ref{fig:naa}. It introduces a brightness-modulated attention sharpening mechanism. For each decoder layer, given the upsampled feature $g$ and the skip-connection feature $F_{out}^{(l)}$, we first compute the standard attention map $\psi \in [0,1]$ as in a conventional attention gate. The sharpening factor and the modulated attention map are then defined in Eq.~\eqref{eq:naa-sharpen}:
\begin{equation}
\label{eq:naa-sharpen}
\begin{aligned}
s &= 1.0 + (1.0 - b) \times 0.5, \\
\psi_{mod} &= \psi^s,
\end{aligned}
\end{equation}
where $s$ is the sharpening factor, which is a linear function of the brightness $b$. In dark scenes where $b$ is small, $s$ becomes larger, which sharpens the attention map, making the attention weights more focused on the relevant features and suppressing the background noise. In well-lit scenes, $s$ is close to 1, so the attention map remains unchanged. This allows the decoder to adaptively adjust its feature selection strength according to the illumination condition.

\subsection{Boundary Prediction and Loss Function}
To further improve the boundary localization accuracy, which is crucial for path planning, we add an auxiliary EdgeHead branch to the decoder. The EdgeHead takes the shallowest decoder feature as input and predicts the probability of each pixel being a road boundary. This auxiliary branch helps the network pay more attention to boundary regions, which are often under-represented in the main segmentation loss.

To keep the optimization objective consistent with the illumination-adaptive design, we further employ an Adaptive Loss Weighting (ALW) scheme. In addition to the main segmentation loss $\mathcal{L}_{seg}$, ALW uses brightness-dependent positive scales $s_{sta}$, $s_{dia}$, and $s_{edge}$ to modulate a semantic transition-aware term $\mathcal{L}_{sta}$, a depth inconsistency-aware term $\mathcal{L}_{dia}$, and the edge supervision term $\mathcal{L}_{edge}$, respectively. Here, $\mathcal{L}_{sta}$ emphasizes semantic transitions near road boundaries, $\mathcal{L}_{dia}$ penalizes inconsistencies between depth-derived geometry and segmentation predictions, and $\mathcal{L}_{edge}$ supervises the auxiliary boundary branch. The overall loss is defined in Eq.~\eqref{eq:alw-loss}:
\begin{equation}
\label{eq:alw-loss}
\begin{aligned}
\mathcal{L} ={}& \mathcal{L}_{seg} + \lambda_{sta}s_{sta}\mathcal{L}_{sta} \\
&\quad + \lambda_{dia}s_{dia}\mathcal{L}_{dia} + \lambda_{edge}s_{edge}\mathcal{L}_{edge},
\end{aligned}
\end{equation}
where $\mathcal{L}_{seg}$ is the cross-entropy loss for road segmentation, $\mathcal{L}_{edge}$ is the boundary prediction loss, and $\lambda_{sta}$, $\lambda_{dia}$, and $\lambda_{edge}$ are the base coefficients of the auxiliary terms. The adaptive scales are predicted from the global brightness estimate $b$ through a lightweight positive mapping function, so that darker scenes can receive stronger auxiliary regularization when needed. In this way, the training objective follows the same illumination cue used in fusion and decoding.

\section{Experiment}
To evaluate the effectiveness of the proposed IAF-Net, we conduct extensive experiments on two datasets: nuScenes-NRS and CARLA-MWRS.

\subsection{Datasets}
\subsubsection{nuScenes Nighttime Road Segmentation (nuScenes-NRS)}
nuScenes-NRS is a large-scale multi-modal autonomous driving dataset derived from nuScenes \cite{nuscenes2020}, which contains a large number of nighttime driving scenes. However, the original dataset does not provide 2D pixel-level road segmentation annotations. In this work, the road segmentation ground truth is generated from the 3D lidar semantic annotations through an automatic projection-and-triangulation pipeline.

The pipeline consists of four steps: (1) Filter the lidar points that belong to the drivable surface class. (2) Project the 3D road points into the image plane using the calibration parameters. (3) Perform Delaunay triangulation on the projected points to fill the sparse points into a dense mask. (4) Filter out invalid connections using a maximum edge length threshold. This automatic annotation setting is related to recent annotation-free drivable-area and curb detection from LiDAR point cloud maps \cite{ma2026annotation}. Using this pipeline, we automatically generate 3182 training images and 805 validation images with pixel-level road annotations, without any manual labeling effort.

\subsubsection{CARLA Multi-Weather Road Segmentation (CARLA-MWRS)}
To evaluate the robustness of our method under extreme weather conditions, we construct CARLA-MWRS using the CARLA simulator \cite{carla2017}. We collect data from three different towns, covering highway, urban, and suburban road scenes. We generate data under four different weather conditions: sunny daytime, sunny nighttime, foggy nighttime, and heavy rain foggy nighttime. This allows us to test the model's performance under various challenging conditions. In total, we collect 3600 images, with 2400 for training and 1200 for testing.

\subsection{Experimental Setup}
We implement our model using PyTorch. The dual-branch encoder is initialized from pre-trained Swin-T weights. Training uses the AdamW optimizer with an initial learning rate of $1\times10^{-4}$ and a weight decay of 0.01. For the main experiments, the batch size is 8. We use a ReduceLROnPlateau scheduler driven by validation performance, gradient clipping with a maximum norm of 1.0, and early stopping after convergence. For efficiency, surface normal inputs are pre-computed from depth in the main training and evaluation pipeline.

For evaluation, we use the standard metrics for road segmentation: MaxF score, IoU (Intersection over Union), Precision, and Recall.

\subsection{Comparison with State-of-the-Art Methods}
We compare our method with six representative road segmentation methods on the nuScenes-NRS validation set. The baseline methods include both RGB-only methods and multi-modal methods. The results are shown in Table~\ref{tab:comparison_nuscenes}.

\begin{table*}[t]
    \centering
    \caption{Performance comparison with representative methods on the nuScenes-NRS validation set. Best results are marked in bold.}
    \label{tab:comparison_nuscenes}
    \resizebox{0.8\textwidth}{!}{
    \begin{tabular}{lcccccc}
        \toprule
        Method & Modality & Params (M) & MaxF (\%) & IoU (\%) & Precision (\%) & Recall (\%) \\
        \midrule
        PIDNet-S \cite{xu2023pidnet} & RGB & 7.6 & 92.43 & 85.92 & 89.83 & 95.18 \\
        DAFormer \cite{hoyer2022daformer} & RGB & 85.2 & 92.58 & 85.66 & 92.55 & 92.00 \\
        DDRNet-23 \cite{hong2021ddrnet} & RGB & 20.3 & 93.23 & 87.26 & 92.51 & 93.96 \\
        SegFormer-B2 \cite{xie2021segformer} & RGB & 27.3 & 93.46 & 87.63 & 92.89 & 94.04 \\
        \midrule
        SNE-RoadSegV2 \cite{feng2025sneroadsegv2} & RGB+N & 73.6 & 95.54 & 91.47 & 95.17 & 95.93 \\
        RoadFormer+ \cite{huang2025roadformerplus} & RGB+N & 131.3 & 95.96 & 92.23 & 96.24 & 95.68 \\
        \midrule
        IAF-Net (Ours) & RGB+N & 73.78 & \textbf{96.11} & \textbf{92.51} & 95.23 & \textbf{97.02} \\
        \bottomrule
    \end{tabular}
    }
\end{table*}

The comparison results lead to three main observations:
1. Multi-modal methods outperform the RGB-only baselines on this benchmark. The best RGB-only method achieves a MaxF score of 93.46\%, while the multi-modal methods all achieve over 95.5\%, suggesting that geometric features provide useful complementary information for nighttime segmentation.
2. IAF-Net achieves the best performance among the compared methods. It improves over the strong baseline RoadFormer+ by 0.15\% in MaxF and 0.28\% in IoU, and over SNE-RoadSegV2 by 0.57\% in MaxF, indicating that illumination-adaptive fusion provides additional benefit beyond standard RGB-normal fusion.
3. Compared with RoadFormer+, our method uses substantially fewer parameters (73.78M vs 131.3M) while still achieving better segmentation performance, indicating a stronger accuracy-efficiency trade-off on this benchmark.

The qualitative results in Fig.~\ref{fig:qualitative_nuscenes} further show that the proposed method preserves the main drivable region even in very dark scenes, with most residual errors concentrated near distant boundaries.

\subsection{Robustness Evaluation on CARLA-MWRS}
To evaluate the robustness of our method under different weather conditions, we test it on CARLA-MWRS. The results are shown in Table~\ref{tab:comparison_carla}.

\begin{table}[t]
\centering
\caption{Performance of IAF-Net under different weather conditions on CARLA-MWRS.}
\label{tab:comparison_carla}
\resizebox{0.9\linewidth}{!}{
    \begin{tabular}{lcc}
        \toprule
        Weather Condition & MaxF (\%) & IoU (\%) \\
        \midrule
        Sunny Daytime & 96.74 & 93.69 \\
        Sunny Nighttime & 97.18 & 94.52 \\
        Foggy Nighttime & 94.19 & 88.92 \\
        Heavy Rain Foggy Nighttime & 94.79 & 90.09 \\
        \midrule
        Overall & 95.73 & 91.81 \\
        \bottomrule
    \end{tabular}
}
\end{table}

The CARLA results show that the proposed method remains stable across all evaluated weather conditions. Under the heavy-rain foggy nighttime condition, it still achieves a MaxF score of 94.79\%. The performance on sunny nighttime is slightly higher than that on sunny daytime, which is likely related to the clearer contrast between road and non-road regions in this synthetic setting.

\subsection{Ablation Study}
To verify the effectiveness of each proposed module with manageable training cost, we conduct ablation studies on nuScenes-NRS using a 40\% training subset with the full validation set. The results are shown in Table~\ref{tab:ablation}.

\begin{table}[t]
    \centering
    \caption{Ablation results on nuScenes-NRS using a 40\% training subset and the full validation set.}
    \label{tab:ablation}
    {\renewcommand{\arraystretch}{1.18}
    \setlength{\tabcolsep}{7pt}
        \begin{tabular}{lccc}
            \toprule
            Configuration & MaxF (\%) & IoU (\%) & Gain \\
            \midrule
            Full IAF-Net (40\%) & 92.91 & 86.77 & - \\
            w/o IAF & 92.21 & 85.55 & +0.70 \\
            w/o NAA & 92.28 & 85.67 & +0.63 \\
            w/o LLEM & 92.55 & 86.13 & +0.36 \\
            w/o ALW & 92.66 & 86.32 & +0.25 \\
            w/o EdgeHead & 92.70 & 86.41 & +0.21 \\
            \midrule
            \makecell[l]{Baseline \\ (R-SNE + backbone)} & 91.62 & 84.54 & +1.29 \\
            \bottomrule
        \end{tabular}
    }
\end{table}

The ablation results show that all proposed modules contribute to the performance improvement under the 40\% training-subset setting. The IAF module provides the largest individual gain, improving MaxF by 0.70\%, while the NAA module provides the second-largest gain of 0.63\%. The LLEM, ALW, and EdgeHead modules also provide consistent improvements. Overall, the full model improves MaxF by 1.29\% over the simplified baseline, supporting the benefit of the complete illumination-adaptive design. The absolute scores in Table~\ref{tab:ablation} are therefore not directly comparable to the full-data comparison in Table~\ref{tab:comparison_nuscenes}.

The ablation results indicate that the proposed modules are complementary, with IAF and NAA providing the largest gains.

\section{Conclusion}
In this paper, we present IAF-Net, an illumination-adaptive fusion network for low-light urban road segmentation. The framework explicitly models illumination conditions and uses this cue to regulate multi-modal fusion and decoding. We also construct two dedicated datasets to facilitate the training and evaluation of low-light road segmentation models. Experiments on the real-world nuScenes-NRS benchmark show state-of-the-art overall performance among the compared methods, while the synthetic CARLA-MWRS benchmark further validates robustness across adverse weather conditions. Ablation studies confirm the importance of the proposed modules, especially the illumination-adaptive fusion design.

Future work will focus on improving efficiency toward real-time deployment and extending the framework to more modalities and more complex scenarios such as rainy and snowy conditions. Another direction is to study tighter integration with downstream planning and control modules.

% \normalem
% \renewcommand{\bibfont}{\small}
% \setlength{\leftmargini}{5pt}
% \setlength{\bibhang}{1.5em}

% \normalem
% \renewcommand{\bibfont}{\small}
% \setlength{\bibhang}{0pt}
% \printbibliography

\defbibenvironment{bibliography}
  {\list
     {\printtext[labelnumberwidth]{%
        \printfield{labelprefix}%
        \printfield{labelnumber}}}
     {\setlength{\labelwidth}{\labelnumberwidth}%
      \setlength{\leftmargin}{\labelwidth}%
      \addtolength{\leftmargin}{\labelsep}%
      \setlength{\itemsep}{\bibitemsep}%
      \setlength{\parsep}{\bibparsep}}%
      \renewcommand*{\makelabel}[1]{##1\hss}}
  {\endlist}
  {\item}

\normalem
\renewcommand{\bibfont}{\footnotesize}
\setlength{\bibitemsep}{0pt}
\setlength{\labelsep}{0.25em}
\printbibliography

@inproceedings{deeplabv3plus2018,
  title={Encoder-Decoder with Atrous Separable Convolution for Semantic Image Segmentation},
  author={Chen, Liang-Chieh and Zhu, Yukun and Papandreou, George and Schroff, Florian and Adam, Hartwig},
  booktitle={Computer Vision -- ECCV 2018},
  pages={833--851},
  year={2018},
  publisher={Springer International Publishing}
}

@inproceedings{pspnet2017,
  title={Pyramid Scene Parsing Network},
  author={Zhao, Hengshuang and Shi, Jianping and Qi, Xiaojuan and Wang, Xiaogang and Jia, Jiaya},
  booktitle={Proceedings of the IEEE Conference on Computer Vision and Pattern Recognition},
  pages={6230--6239},
  year={2017},
}

@inproceedings{xie2021segformer,
  title={{SegFormer}: Simple and Efficient Design for Semantic Segmentation with Transformers},
  author={Xie, Enze and Wang, Wenhai and Yu, Zhiding and Anandkumar, Anima and Alvarez, Jose M. and Luo, Ping},
  booktitle={Advances in Neural Information Processing Systems},
  volume={34},
  pages={12077--12090},
  year={2021}
}

@inproceedings{xu2023pidnet,
  title={{PIDNet}: A Real-Time Semantic Segmentation Network Inspired by {PID} Controllers},
  author={Xu, Jiacong and Xiong, Zixiang and Bhattacharyya, Shankar P.},
  booktitle={Proceedings of the IEEE/CVF Conference on Computer Vision and Pattern Recognition},
  pages={19529--19539},
  year={2023}
}

@article{hong2021ddrnet,
  title={Deep Dual-Resolution Networks for Real-Time and Accurate Semantic Segmentation of Traffic Scenes},
  author={Pan, Huihui and Hong, Yuanduo and Sun, Weichao and Jia, Yisong},
  journal={IEEE Trans. Intell. Transp. Syst.},
  volume={24},
  number={3},
  pages={3448--3460},
  year={2023},
}

@article{retinex1971,
  title={Lightness and Retinex Theory},
  author={Land, Edwin H and McCann, John J},
  journal={J. Opt. Soc. Am.},
  volume={61},
  number={1},
  pages={1--11},
  year={1971},
  publisher={Optica Publishing Group}
}

@inproceedings{kind2019,
  title={Kindling the Darkness: A Practical Low-Light Image Enhancer},
  author={Zhang, Yonghua and Zhang, Jiawan and Guo, Xiaojie},
  booktitle={Proceedings of the 27th ACM International Conference on Multimedia},
  pages={1632--1640},
  year={2019},
}

@article{llnet2015,
  title={{LLNet}: A Deep Autoencoder Approach to Natural Low-Light Image Enhancement},
  author={Lore, Kin Gwn and Akintayo, Adedotun and Sarkar, Soumik},
  journal={Pattern Recognition},
  volume={61},
  pages={650--662},
  year={2017},
  publisher={Elsevier}
}

@inproceedings{wei2023disentangle,
  title={Disentangle then Parse: Night-time Semantic Segmentation with Illumination Disentanglement},
  author={Wei, Zhixiang and Chen, Lin and Tu, Tao and Ling, Pengyang and Chen, Huaian and Jin, Yi},
  booktitle={Proceedings of the IEEE/CVF International Conference on Computer Vision},
  pages={21536--21546},
  year={2023},
}

@inproceedings{heatnet2019,
  title={HeatNet: Bridging the Day-Night Domain Gap in Semantic Segmentation with Thermal Images},
  author={Vertens, Johan and Z{\"u}rn, Jannik and Burgard, Wolfram},
  booktitle={2020 IEEE/RSJ International Conference on Intelligent Robots and Systems (IROS)},
  pages={8461--8468},
  year={2020},
  organization={IEEE}
}

@article{transnightseg2024,
  title={Semantic segmentation method on nighttime road scene based on {Trans-nightSeg}},
  author={Li, Canlin and Zhang, Wenjiao and Shao, Zhiwen and Ma, Lizhuang and Wang, Xinyue},
  journal={Journal of Zhejiang University (Engineering Science)},
  volume={58},
  number={2},
  pages={294--303},
  year={2024},
}

@article{zhang2023cmx,
  title={{CMX}: Cross-Modal Fusion for {RGB-X} Semantic Segmentation With Transformers},
  author={Zhang, Jiaming and Liu, Huayao and Yang, Kailun and Hu, Xinxin and Liu, Ruiping and Stiefelhagen, Rainer},
  journal={IEEE Trans. Intell. Transp. Syst.},
  volume={24},
  number={12},
  pages={14679--14694},
  year={2023},
}

@article{broedermann2025cafuser,
  title={CAFuser: Condition-Aware Multimodal Fusion for Robust Semantic Perception of Driving Scenes},
  author={Br{\"o}dermann, Tim and Sakaridis, Christos and Fu, Yuqian and Van Gool, Luc},
  journal={IEEE Robot. Autom. Lett.},
  volume={10},
  number={4},
  pages={3134--3141},
  year={2025},
}

@article{li2024roadformer,
  title={RoadFormer: Duplex Transformer for RGB-Normal Semantic Road Scene Parsing},
  author={Li, Jiahang and Zhang, Yikang and Yun, Peng and Zhou, Guangliang and Chen, Qijun and Fan, Rui},
  journal={IEEE Trans. Intell. Veh.},
  volume={9},
  number={7},
  pages={5163--5172},
  year={2024},
}

@article{huang2025roadformerplus,
  title={RoadFormer+: Delivering {RGB-X} Scene Parsing Through Scale-Aware Information Decoupling and Advanced Heterogeneous Feature Fusion},
  author={Huang, Jianxin and Li, Jiahang and Jia, Ning and Sun, Yuxiang and Liu, Chengju and Chen, Qijun and Fan, Rui},
  journal={IEEE Trans. Intell. Veh.},
  volume={10},
  number={5},
  pages={3156--3165},
  year={2025},
}

@inproceedings{swin2021,
  title={Swin Transformer: Hierarchical Vision Transformer Using Shifted Windows},
  author={Liu, Ze and Lin, Yutong and Cao, Yue and Hu, Han and Wei, Yixuan and Zhang, Zheng and Lin, Stephen and Guo, Baining},
  booktitle={Proceedings of the IEEE/CVF International Conference on Computer Vision},
  pages={9992--10002},
  year={2021},
}

@inproceedings{oktay2018attention,
  title={Attention {U-Net}: Learning Where to Look for the Pancreas},
  author={Oktay, Ozan and Schlemper, Jo and Folgoc, Loic Le and Lee, Matthew and Heinrich, Mattias and Misawa, Kazunari and Mori, Kensaku and McDonagh, Steven and Hammerla, Nils Y. and Kainz, Bernhard and Glocker, Ben and Rueckert, Daniel},
  booktitle={Medical Imaging with Deep Learning},
  year={2018}
}

@inproceedings{nuscenes2020,
  title={{nuScenes}: A Multimodal Dataset for Autonomous Driving},
  author={Caesar, Holger and Bankiti, Varun and Lang, Alex H. and Vora, Sourabh and Liong, Venice Erin and Xu, Qiang and Krishnan, Anush and Pan, Yu and Baldan, Giancarlo and Beijbom, Oscar},
  booktitle={Proceedings of the IEEE/CVF Conference on Computer Vision and Pattern Recognition},
  pages={11618--11628},
  year={2020},
}

@inproceedings{carla2017,
  title={{CARLA}: An Open Urban Driving Simulator},
  author={Dosovitskiy, Alexey and Ros, German and Codevilla, Felipe and Lopez, Antonio and Koltun, Vladlen},
  booktitle={Proceedings of the 1st Annual Conference on Robot Learning},
  pages={1--16},
  year={2017},
  volume={78},
  series={Proceedings of Machine Learning Research},
  publisher={PMLR}
}

@inproceedings{hoyer2022daformer,
  title={{DAFormer}: Improving Network Architectures and Training Strategies for Domain-Adaptive Semantic Segmentation},
  author={Hoyer, Lukas and Dai, Dengxin and Van Gool, Luc},
  booktitle={Proceedings of the IEEE/CVF Conference on Computer Vision and Pattern Recognition},
  pages={9914--9925},
  year={2022},
}

@article{feng2025sneroadsegv2,
  title={SNE-RoadSegV2: Advancing Heterogeneous Feature Fusion and Fallibility Awareness for Freespace Detection},
  author={Feng, Yi and Ma, Yu and Andreev, Stepan and Chen, Qijun and Dvorkovich, Alexander and Pitas, Ioannis and Fan, Rui},
  journal={IEEE Trans. Instrum. Meas.},
  volume={74},
  pages={1--9},
  year={2025},
}

@article{ghari2024lowlightsurvey,
  title={Pedestrian Detection in Low-Light Conditions: A Comprehensive Survey},
  author={Ghari, Bahareh and Tourani, Ali and Shahbahrami, Asadollah and Gaydadjiev, Georgi},
  journal={Image and Vision Computing},
  volume={148},
  pages={105106},
  year={2024},
}

@inproceedings{ma2023self,
  title={Self-Supervised Drivable Area Segmentation Using {LiDAR}'s Depth Information for Autonomous Driving},
  author={Ma, Fulong and Liu, Yang and Wang, Sheng and Wu, Jin and Qi, Weiqing and Liu, Ming},
  booktitle={2023 IEEE/RSJ International Conference on Intelligent Robots and Systems (IROS)},
  pages={41--48},
  year={2023},
  organization={IEEE}
}

@misc{ma2026annotation,
  title={Annotation-Free Detection of Drivable Areas and Curbs Leveraging LiDAR Point Cloud Maps},
  author={Ma, Fulong and Peng, Daojie and Ma, Jun},
  year={2026},
  eprint={2603.27553},
  archivePrefix={arXiv},
  primaryClass={cs.CV}
}

@article{chen2019progressive,
  title={Progressive {LiDAR} Adaptation for Road Detection},
  author={Chen, Zhe and Zhang, Jing and Tao, Dacheng},
  journal={IEEE/CAA J. Autom. Sinica},
  volume={6},
  number={3},
  pages={693--702},
  year={2019},
  publisher={IEEE}
}

@inproceedings{sun2025rod,
  title={{ROD}: {RGB}-Only Fast and Efficient Off-Road Freespace Detection},
  author={Sun, Tong and Ye, Hongliang and Mei, Jilin and Chen, Liang and Zhao, Fangzhou and Zong, Leiqiang and Hu, Yu},
  booktitle={2025 IEEE International Conference on Robotics and Automation (ICRA)},
  pages={9787--9793},
  year={2025},
  organization={IEEE}
}

@inproceedings{milioto2019rangenet++,
  title={{RangeNet++}: Fast and Accurate {LiDAR} Semantic Segmentation},
  author={Milioto, Andres and Vizzo, Ignacio and Behley, Jens and Stachniss, Cyrill},
  booktitle={2019 IEEE/RSJ International Conference on Intelligent Robots and Systems (IROS)},
  pages={4213--4220},
  year={2019},
  organization={IEEE}
}

@inproceedings{min2022orfd,
  title={{ORFD}: A Dataset and Benchmark for Off-Road Freespace Detection},
  author={Min, Chen and Jiang, Weizhong and Zhao, Dawei and Xu, Jiaolong and Xiao, Liang and Nie, Yiming and Dai, Bin},
  booktitle={2022 International Conference on Robotics and Automation (ICRA)},
  pages={2532--2538},
  year={2022},
  organization={IEEE}
}

@inproceedings{zerodce2020,
  title={Zero-Reference Deep Curve Estimation for Low-Light Image Enhancement},
  author={Guo, Chunle and Li, Chongyi and Guo, Jichang and Loy, Chen Change and Hou, Junhui and Kwong, Sam and Cong, Runmin},
  booktitle={2020 IEEE/CVF Conference on Computer Vision and Pattern Recognition (CVPR)},
  pages={1777--1786},
  year={2020},
  organization={IEEE}
}

@inproceedings{dannet2021,
  title={{DANNet}: A One-Stage Domain Adaptation Network for Unsupervised Nighttime Semantic Segmentation},
  author={Wu, Xinyi and Wu, Zhenyao and Guo, Hao and Ju, Lili and Wang, Song},
  booktitle={2021 IEEE/CVF Conference on Computer Vision and Pattern Recognition (CVPR)},
  pages={15764--15773},
  year={2021},
  organization={IEEE}
}

@inproceedings{ccdistill2022,
  title={Cross-Domain Correlation Distillation for Unsupervised Domain Adaptation in Nighttime Semantic Segmentation},
  author={Gao, Huan and Guo, Jichang and Wang, Guoli and Zhang, Qian},
  booktitle={2022 IEEE/CVF Conference on Computer Vision and Pattern Recognition (CVPR)},
  pages={9903--9913},
  year={2022},
  organization={IEEE}
}

@inproceedings{cmda2023,
  title={{CMDA}: Cross-Modality Domain Adaptation for Nighttime Semantic Segmentation},
  author={Xia, Ruihao and Zhao, Chaoqiang and Zheng, Meng and Wu, Ziyan and Sun, Qiyu and Tang, Yang},
  booktitle={2023 IEEE/CVF International Conference on Computer Vision (ICCV)},
  pages={21515--21524},
  year={2023},
  organization={IEEE}
}

@article{eventaseg2024,
  title={{EventASEG}: An Event-Based Asynchronous Segmentation of Road With Likelihood Attention},
  author={Annamalai, Lakshmi and Thakur, Chetan Singh},
  journal={IEEE Robot. Autom. Lett.},
  volume={9},
  number={8},
  pages={6951--6958},
  year={2024}
}

@article{wang2019raldrivable,
  title={Self-Supervised Drivable Area and Road Anomaly Segmentation Using {RGB-D} Data for Robotic Wheelchairs},
  author={Wang, Hengli and Sun, Yuxiang and Liu, Ming},
  journal={IEEE Robot. Autom. Lett.},
  volume={4},
  number={4},
  pages={4386--4393},
  year={2019}
}

@article{zheng2023robustperception,
  title={Robust Perception Under Adverse Conditions for Autonomous Driving Based on Data Augmentation},
  author={Zheng, Ziqiang and Cheng, Yujie and Xin, Zhichao and Yu, Zhibin and Zheng, Bing},
  journal={IEEE Trans. Intell. Transp. Syst.},
  volume={24},
  number={12},
  pages={13916--13929},
  year={2023}
}

@article{peng2026litevilnet,
  title={LiteViLNet: Lightweight Vision-LiDAR Fusion Network for Efficient Road Segmentation},
  author={Peng, Daojie and Wang, Bingtao and Ma, Fulong and Zhang, Liang and Ma, Jun},
  journal={arXiv preprint arXiv:2605.21007},
  year={2026}
}

\end{document}